\documentclass[conference]{IEEEtran}
\IEEEoverridecommandlockouts
\usepackage{cite}
\usepackage{amsmath,amssymb,amsfonts}
\usepackage{algorithmic}
\usepackage{graphicx}
\usepackage{textcomp}
\usepackage{xcolor}
\def\BibTeX{{\rm B\kern-.05em{\sc i\kern-.025em b}\kern-.08em
    T\kern-.1667em\lower.7ex\hbox{E}\kern-.125emX}}
    
\usepackage{makecell}
\usepackage{amsmath}
\usepackage{bm}
\usepackage{graphicx, caption, subcaption}

\definecolor{citecolor}{HTML}{0071bc}
\usepackage[colorlinks=true,linkcolor=red, citecolor=citecolor]{hyperref}  

\title{Incremental Learning with\\Differentiable Architecture and Forgetting Search}

\iftrue
    \author{
    \IEEEauthorblockN{James Seale Smith}
    \IEEEauthorblockA{
    \textit{Georgia Institute of Technology}\\
    Atlanta, Georgia, USA \\
    jamessealesmith@gatech.edu}
    \and
    \IEEEauthorblockN{Zachary Seymour}
    \IEEEauthorblockA{
    \textit{SRI International}\\
    Princeton, NJ, USA \\
    yzachary.seymour@sri.com}
    \and
    \IEEEauthorblockN{Han-Pang Chiu}
    \IEEEauthorblockA{
    \textit{SRI International}\\
    Princeton, NJ, USA \\
    han-pang.chiu@sri.com}
    }
\else
     \author{\IEEEauthorblockN{Anonymous Authors}}
\fi

\begin{document}

\maketitle

\begin{abstract}
    As progress is made on training machine learning models on incrementally expanding classification tasks (i.e., incremental learning), a next step is to translate this progress to industry expectations. One technique missing from incremental learning is automatic architecture design via Neural Architecture Search (NAS). In this paper, we show that leveraging NAS for incremental learning results in strong performance gains for classification tasks. Specifically, we contribute the following: first, we create a strong baseline approach for incremental learning based on Differentiable Architecture Search (DARTS) and state-of-the-art incremental learning strategies, outperforming many existing strategies trained with similar-sized popular architectures; second, we extend the idea of architecture search to regularize architecture forgetting, boosting performance past our proposed baseline. We evaluate our method on both RF signal and image classification tasks, and demonstrate we can achieve up to a 10\% performance increase over state-of-the-art methods. Most importantly, our contribution enables learning from continuous distributions on real-world application data for which the complexity of the data distribution is unknown, or the modality less explored (such as RF signal classification).
\end{abstract}


\begin{figure*}[t]
    \centering
    \begin{subfigure}[t]{0.62\textwidth}
        \centering
        \includegraphics[trim=0 0 0 0, clip, width=300pt]{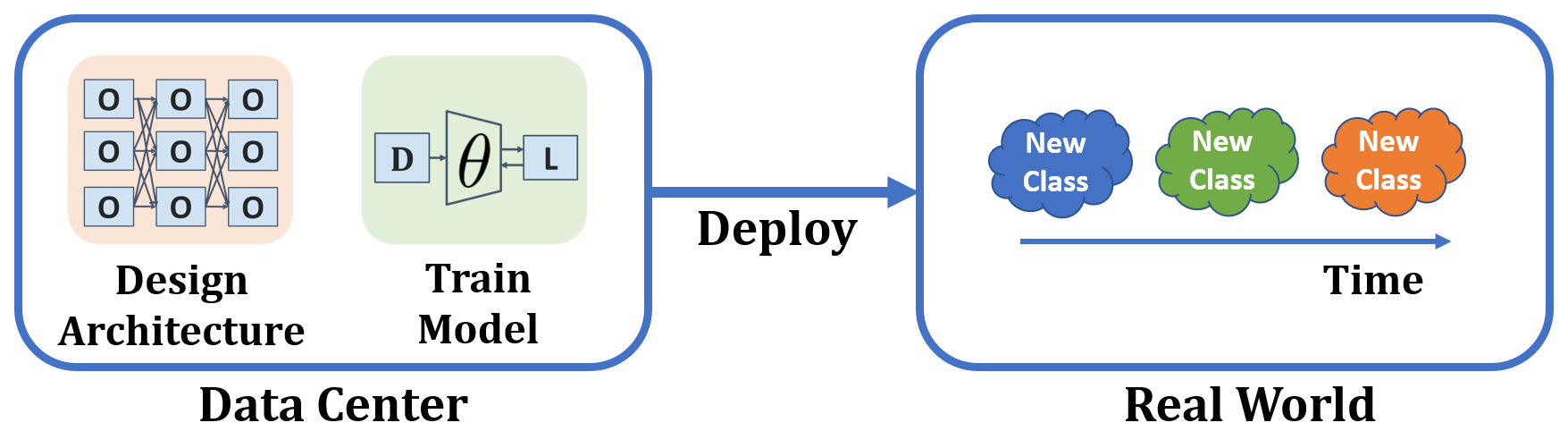}
        \vspace{-3mm}
        \caption{}
        \vspace{1mm}
        \includegraphics[trim=0 0 0 0, clip, width=300pt]{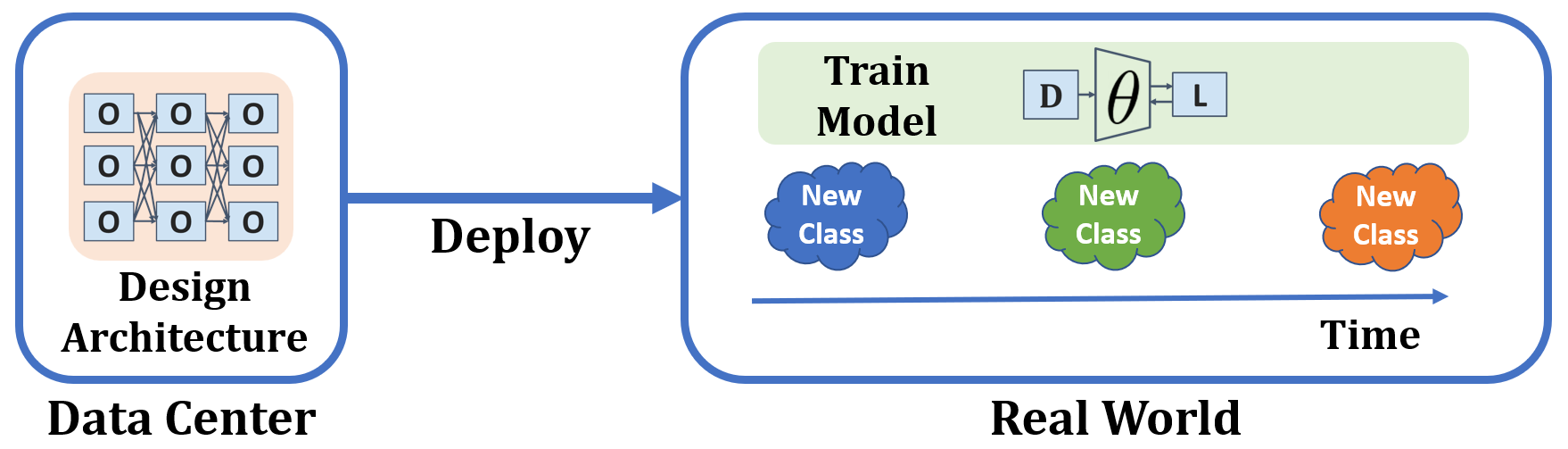}
        \vspace{-3mm}
        \caption{}
    \end{subfigure}
    \hfill
    \begin{subfigure}[h]{0.37\textwidth}
        \centering
        \vspace{15mm}
        \includegraphics[trim=0 0 0 0, clip, width=180pt]{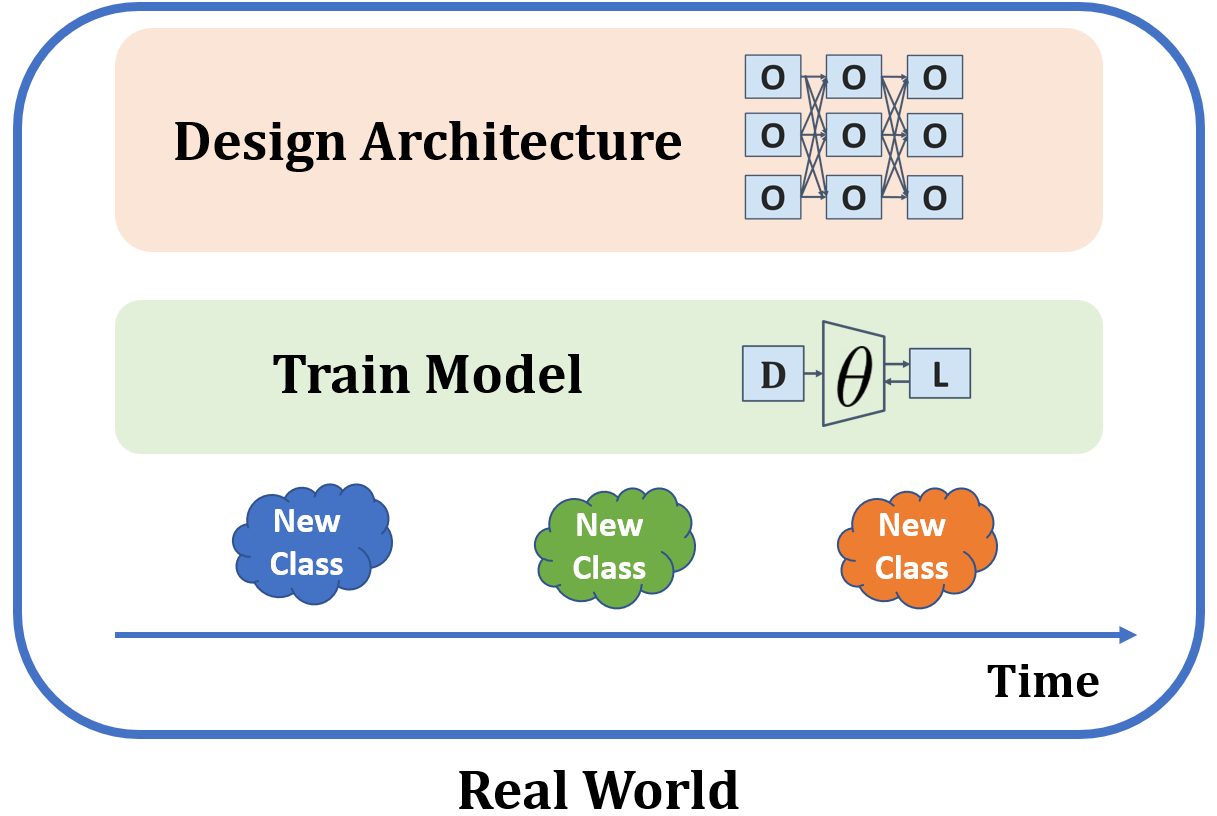}
        \vspace{-1mm}
        \caption{}
    \end{subfigure}
    \caption{(a) A typical machine learning model has both its architecture designed and weights trained in an offline manner at a data center. After deployment, this model will suffer when seeing new classes in the real world. (b) An incremental learning model has its architecture designed at a data center, but will update its model weights after deployment to incorporate new classes in the real world. (c) We learn \text{both} the architecture \text{and} weights after deployment, which is crucial for incremental learning for data in less explored modalities or of unknown underlying complexity.}
	\label{fig:main-idea}
\end{figure*}

\section{Introduction}

Machine learning applications built on deep neural networks are typically trained offline on a large dataset. The model is frozen and deployed, and it assumes the distribution of data it will see is of the same distribution of the training dataset. Unfortunately, this is not the case for many real-world applications. The distribution of seen data may have a \textit{domain shift}, leading to prediction errors from the model.

This \textit{domain shift} may come in the form of the \textit{pixel space} (e.g., different weather conditions affecting background and lighting) or the \textit{semantic space} (e.g., we encounter novel objects which were not seen in the training data). These shifts are typically not a single instance; often, the distribution of data continuously shifts over the lifetime of deployed application, meaning we must continuously update the model in response to these shifts. In this paper, we focus on \textit{continuous domain shifts} in the \textit{semantic space}, a paradigm commonly referred to as \textit{class-incremental continual learning}, or incremental learning for short.

Models designed for the incremental learning setting should be able to incorporate new information about novel classes seen in a continuously shifting datastream, while simultaneously avoiding \textit{catastrophic forgetting} of previously seen classes which may disappear from the stream for extended periods of time~\cite{goodfellow2013empirical}. Specifically, the broad literature in typical incremental learning asks us: ``how can we acquire new semantic classes without forgetting previously seen classes?"

Another concurrent challenge in machine learning applications is selecting an appropriate neural architecture for data of unknown complexities or less-explored modality. For example, while the commonly used ResNet architecture~\cite{he2016deep} achieves high performance on image datasets such as ImageNet~\cite{russakovsky2015imagenet}, we can often achieve higher performance at a lower cost in inference time and model parameters by using a neural architecture selected or learned in an automated fashion. Additionally, architectures have been less explored for modalities such as RF signal classification~\cite{o2018over}, giving us fewer pre-existing options to explore. We have motivated and described the paradigm of \textit{Neural Architecture Search} (NAS), a broad research area which pushes to automatically find neural architectures to boost model performance for real-world deep learning applications.

Visualized in Figure~\ref{fig:main-idea}, we unify the two crucial concepts of \textit{incremental learning} and \textit{neural architecture search} to design a flexible incremental learning approach which can be applied to semantically-shifting data of unknown underlying complexity. This could be important for applications such as on-device user-specific learning (where the complexity of user actions may be unknown), exploratory agents relying on non-pixel image perception (where novel objects are expected), or animal motion detection with (where it may be cheap to rely on dated, pre-existing sensors for which finding a neural architecture may be difficult).

Our high level approach is to first integrate state-of-the-art incremental learning strategies (including replay, knowledge distillation regularization, class-balanced fine-tuning, etc.)~\cite{castro2018end} into a popular and efficient NAS method, Differentiable Architecture Search (DARTS). This required a combination of high-level design and numerous empirical analyses, and resulted in up to a 10\% increase in final accuracy compared to the state-of-the-art. We took our work one step forward, using analysis of the inferred architecture at different tasks to motivate a regularization penalty on the architecture decisions themselves, resulting in additional performance gains. In summary, we make the following contributions:

\begin{enumerate}
    \item We unify the concepts of \textit{incremental learning} and \textit{neural architecture search} to propose learning from continuous distributions of data from less explored modalities or unknown underlying complexity. 
    \item We propose Incremental-DARTS (I-DARTS), which combines SOTA incremental learning strategies with the DARTS method to increase incremental classification accuracy by as much as 10\%.
    \item We use analysis of the I-DARTS architecture over tasks to motivate an architecture decision regularization, resulting in additional performance gains.
\end{enumerate}

\section{Background and Related Work}

\textbf{Catastrophic Forgetting}: Methods to mitigate catastrophic forgetting can broadly be separated into either (i) architecture expansion or (ii) regularization. Architecture expansion~\cite{ebrahimi2020adversarial,lee2020neural,lomonaco2017core50,maltoni2019continuous,Rusu:2016} is useful for applications in which the architecture can grow in parameters with the number of tasks. However, these are typically associated with \textit{task-incremental learning} rather than \textit{class-incremental learning}, an important nuance (discussed in the next section) which cannot be compared to our setting. Regularization approaches mitigate forgetting by regularizing the model with respect to past task knowledge while training on a new new task. This can be done by either regularizing the model in the  weight space (i.e., penalize changes to model parameters)~\cite{aljundi2017memory,ebrahimi2019uncertainty,kirkpatrick2017overcoming,titsias2019functional,zenke2017continual} or the prediction space (i.e., penalize changes to model predictions)~\cite{castro2018end,hou2018lifelong,lee2019overcoming,li2016learning}. 
    
\textbf{Neural Architecture Search}: Orthogonal to the catastrophic forgetting problem is concept of Neural Architecture Search (NAS). This type of work automates the process of architecture selection, replacing the expensive (with respect to time and costly errors) engineering of neural architectures with an optimization problem. Because incremental learning (the main focus of this paper) typically associates a cost to training time, our work is only concerned with efficient, one-shot NAS approaches. For more details on other types of NAS, we refer the reader to recent literature surveys~\cite{elsken2019neural,ren2020comprehensive,wistuba2019survey}.

The high level idea of one-shot NAS is to model the space of candidate architectures with a single, large model, and then search for an optimal sub-model within the large ``super-model". While several works~\cite{cai2018proxylessnas,guo2020single,pham2018efficient,wu2019fbnet} fit under this constraint, we selected the DARTS method~\cite{liu2018darts} because (i) its two-step differentiable algorithm fits nicely with existing incremental learning concepts, and (ii) there exists a user-friendly implementation~\cite{nni}, allowing us to focus our work on the incremental learning contributions.

\textbf{Continuous Neural Architecture Search}: Our work is not the first to consider concepts of neural architecture search in a continuous learning setting. Firefly neural architecture search~\cite{wu2021firefly} continuously grows a model for continual learning by searching for the optimal growing method in each task (split existing neurons, grow new neurons, or grow new layers). Learn-to-Grow~\cite{li2019learn} and NSAS~\cite{zhang2020one} also formulates continuous neural architecture search from a parameter reuse perspective. However, these works focus on the \textit{task-incremental learning}, which cannot be compared with our work on \textit{class-incremental learning}. Continually Neural Architecture Search (CNAS)~\cite{huang2019neural} is the first work to propose neural architecture search for class-incremental learning. However, this work assumes access to past task data, with its contributions focusing on the efficient expansion of an architecture rather than considering important continual learning constraints (and we therefore cannot compare to this method). Finally, recent work on neural architecture search of deep priors~\cite{mundt2021neural} has invoking findings on learning a deep prior on which to train a linear classifier without forgetting, but the problem setting assumes training on classes from a similar distribution from the target distribution (which we strongly do \textit{not} assume, given our work is designed specifically for the case of unknown data complexities). \textit{Importantly, none of the discussed works unify the typical incremental learning setting with NAS, a key contribution of our work.}
\section{Preliminaries}

In incremental learning, we show a model labeled data corresponding to $M$ semantic object classes $c_1, c_2, \dots, c_M$ over a series of $N$ tasks corresponding to non-overlapping subsets of classes. We use the notation $\mathcal{T}_n$ to denote the set of classes introduced in task $n$, with $|\mathcal{T}_n|$ denoting the number of object classes in task $n$. To describe our inference model, we denote $\theta_{k,n}$ as the model $\theta$ at time $k$ that has been trained with the classes from task $n$. For example, $\theta_{n,1:n}$ refers to the model trained during task $n$ and its logits associated with all tasks up to and including class $n$. We drop the second index when describing the model trained during task $n$ with all logits (for example, $\theta_{n}$).

In this setting, each class appears in only a single task, and the goal is to incrementally learn to classify new object classes as they are introduced while retaining performance on previously learned classes. The incremental learning setting is challenging because no task indexes are provided to the learner during inference and the learner must support classification across all classes seen up to task $n$~\cite{hsu2018re}. Incremental learning, sometimes referred to as \textit{class-incremental learning}, is more difficult than \textit{task-incremental learning}, where the task indexes are given during both training and inference (which is an entirely different learning setting).

A key component of the incremental learning setting discussed in this work is that competitive methods must sample a small coreset of training data from past tasks~\cite{aljundi2019online, aljundi2019gradient, chaudhry2018efficient,chaudhry2019episodic,Gepperth:2017,hayes2018memory,hou2019learning,Kemker:2017,Lopez-Paz:2017,Rebuffi:2016,robins1995catastrophic, rolnick2019experience,von2019continual}. We denote this coreset as $\mathcal{X}_{core}$, and it will be replayed in future tasks $\mathcal{T}_n$ along with the corresponding training data, denoted as $\mathcal{X}_{n}$.

\begin{figure*}[t!]
    \centering
    \includegraphics[trim=130 90 140 150, clip, height=180pt]{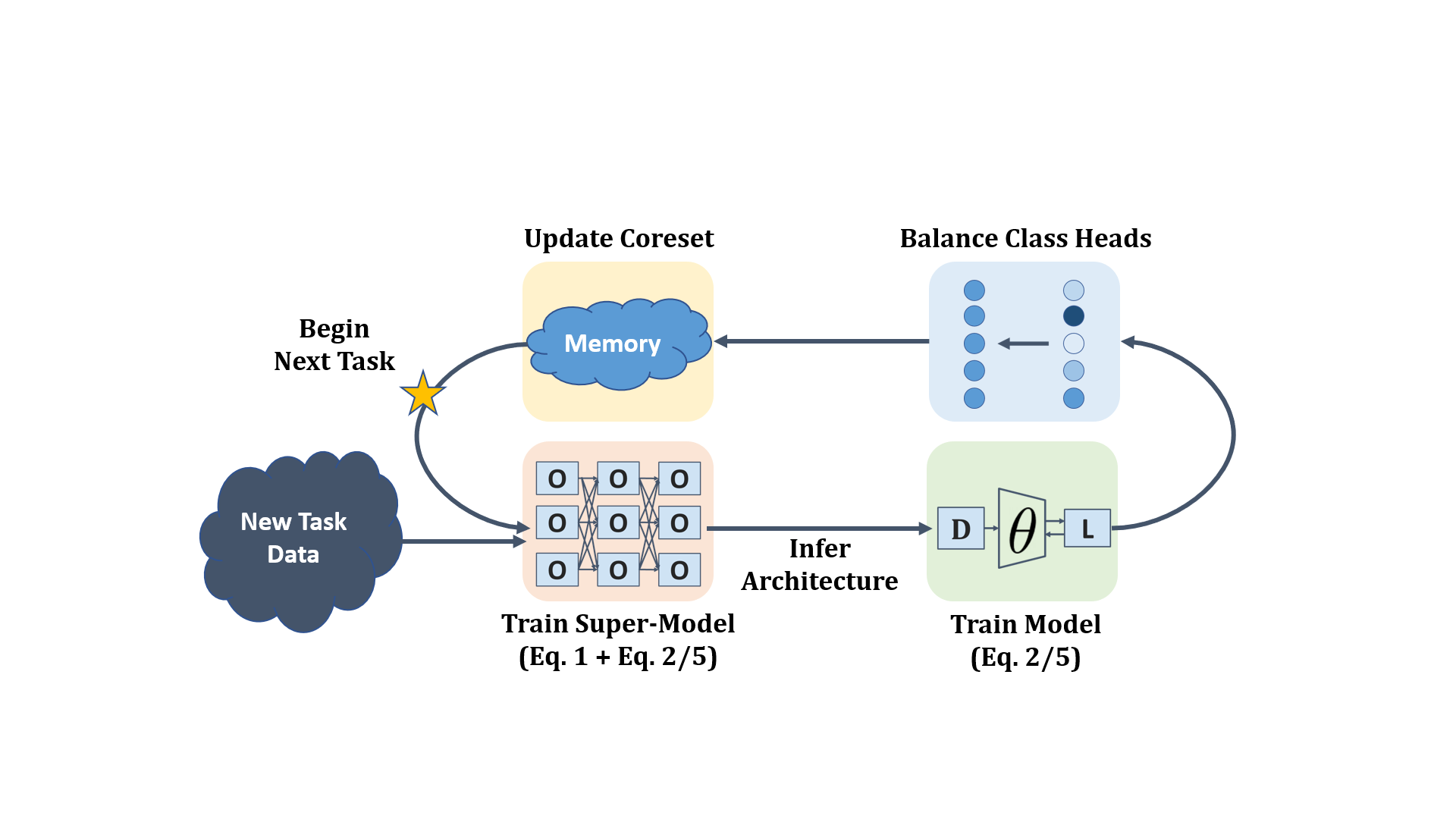}
	\caption{\textbf{Incremental DARTS (I-DARTS).} At each task, we first train our super-model using the bilevel DARTS optimization of Eq. \eqref{eq:darts} using the loss functions given in Eq. \eqref{eq:lwf} (or Eq. \eqref{eq:lwf_alpha} for I-DARTS*). We then infer our optimal neural architecture, and retrain this architecture using all training and coreset data (D) with the same loss functions (L) as the previous step. Next, we apply a class-balancing fine-tuning stage to remove bias in the classification heads. Finally, we update our coreset to best represent the prior task training data. We move to the next task, and repeat this cycle until all tasks have been visited.}
	\label{fig:approach}
\end{figure*}

\section{A Strong Baseline for Incremental Learning}

In this section, we first review the pertinent details of the neural architecture approach, DARTS. As discussed in the previous section, we chose to build our method using the DARTS approach due to its differentiability and ease of implementation. We then describe how we integrate incremental learning strategies into DARTS to create our Incremental-DARTS baseline approach.

\subsection{Differentiable Architecture Search (DARTS)}

The goal of DARTS~\cite{liu2018darts} is to find an optimal neural architecture given a ``super-model" of candidate operations. Specifically, we consider a set of candidate operations $\mathcal{O}$ (e.g., convolution, pooling, identity, \textit{zero}) as our super-model. These operations can be considered as nodes in a directed acyclic graph. At each node, we refer to the intermediate representation as $x^{(i)}$, with each directed edge $(i,j)$ being associated with some operation $o^{(i,j)}$ from $\mathcal{O}$. The goal of DARTS is select which operations $o^{(i,j)}$ at each $(i,j)$ should be used for optimal model performance.

In DARTS, we model the operation at each node as a mixture of the candidate operations at that node. That is, $x^{(i)}$ is transformed in the super-model at each $(i,j)$ by a weighted sum of each $o^{(i,j)}$ transformation. The mixture weights at node $(i,j)$ are parameterized with a vector $\alpha^{(i,j)}$ of dimension $|\mathcal{O}|$. With this formulation, we can now solve a bilevel optimization which iterates optimizing \textit{architecture weights} $w$ (which parameterize the candidate operations) with respect to training data and the \textit{mixture weights} $\alpha$ (which parameterize the weighting of the candidate operations) with respect to holdout data. Formally, we have:

\begin{equation}
\begin{split}
    &\underset{\alpha}{min} \hspace{0.5em} \mathcal{L}_{val} \left(w^*(\alpha),\alpha \right) \\
    &s.t. \quad w^*(\alpha) = \underset{w}{argmin} \hspace{0.5em} \mathcal{L}_{train}(w,\alpha)
\end{split}
\label{eq:darts}
\end{equation}

At the end of this training, we can then infer a discrete architecture using the argmax of the $\alpha$ (i.e., only the $o^{(i,j)}$ at each $(i,j)$ with the highest corresponding $\alpha^{(i,j)}$ is retained). Finally, we retrain the inferred architecture using the combined training and holdout data.

To be consistent with our notation described in the previous section, we will simply refer to model optimization in the rest of this paper as optimizing with respect to $\theta_n$. If we optimize with respect to $\theta_n$ during the bilevel optimization stage of our approach, we will specifically update both architecture weights $w$ and mixture weights $\alpha$. Otherwise, we will only be updating architecture weights $w$ of the inferred neural architecture.
\subsection{Incremental DARTS (I-DARTS)}

An intuitive approach for an incremental variant of DARTS might be to simply run the DARTS algorithm including a coreset of exemplar data for replay. However, this would not take advantage of the numerous advancements in incremental learning which far exceed simple replay data. Here, we propose a strong incremental learning variant of DARTS, referred to as I-DARTS.

We first address catastrophic forgetting with a prediction space regularization, which encourages the model to learn the classes of task $\mathcal{T}_n$ without unlearning the representation of classes of tasks $\mathcal{T}_1 \cdots \mathcal{T}_{n-1}$. We observe here that we incorporate prediction space regularization over model space regularization because the former has been found to perform better for class-incremental learning~\cite{lesort2019generative,van2018generative}. 

Specifically, in task $\mathcal{T}_n$ we include a \textit{knowledge distillation} (KD) loss~\cite{li2016learning} over all data $\mathcal{X}_{core} \cup \mathcal{X}_{n}$, which forces $\theta_n$ to learn $\mathcal{T}_n$ with minimal degradation to $\mathcal{T}_1 \cdots \mathcal{T}_{n-1}$ knowledge. Denoting $p_{\theta_n}(x)$ as the predicted class distribution produced by model $\theta_n$ for some input $x$, we replace the loss $\mathcal{L}$ in Eq.\eqref{eq:darts} to instead minimize: 
\begin{equation}
    \underset{\theta_n}{min} \hspace{0.5em} \mathcal{L}_{CE} \left( p_{\theta_{n,1:n}}(x), y \right) + \mu \mathcal{L}_{KD}(x,\theta_n,\theta_{n-1})
\label{eq:lwf}
\end{equation}
where $\theta_{n-1}$ is a frozen copy of $\theta$ from the end of task $\mathcal{T}_{n-1}$, $\mathcal{L}_{CE}$ is standard cross entropy loss using labels $y$, $\mu$ is a scalar hyperparameter which weights the contribution of regularization penalty, and $\mathcal{L}_{KD}$ is our knowledge distillation regularization given as the KL divergence between two predictions:
\begin{equation}
    \mathcal{L}_{KD}(x,\theta_n,\theta_{n-1}) = KL(p_{\theta_{n-1,1:n-1}}(x) || p_{\theta_{n,1:n-1}}(x)) 
    \label{eq:kd}
\end{equation}

Rather than randomly sampling data for replay, we use a \textit{herding selection}~\cite{welling2009herding}, which has been found to boost performance for incremental learning~\cite{castro2018end}. At a high level, this strategy encourages replay data sampled \textit{for each class} to well resemble the feature distribution of training data in that class. We also include a class-balanced fine-tuning stage~\cite{castro2018end} in our approach to deal with the class-bias balancing issue in incremental learning (a commonly studied problem where models are biased towards current task classes). This stage simply involves fine-tuning the classification heads using only the coreset data (which is balanced by class). The final I-DARTS approach is visualized in Figure~\ref{fig:approach}.
\begin{figure*}[t!]
    \centering
    \begin{subfigure}[t]{0.38\textwidth}
        \centering
        \includegraphics[trim=0 0 0 38, clip, width=\textwidth]{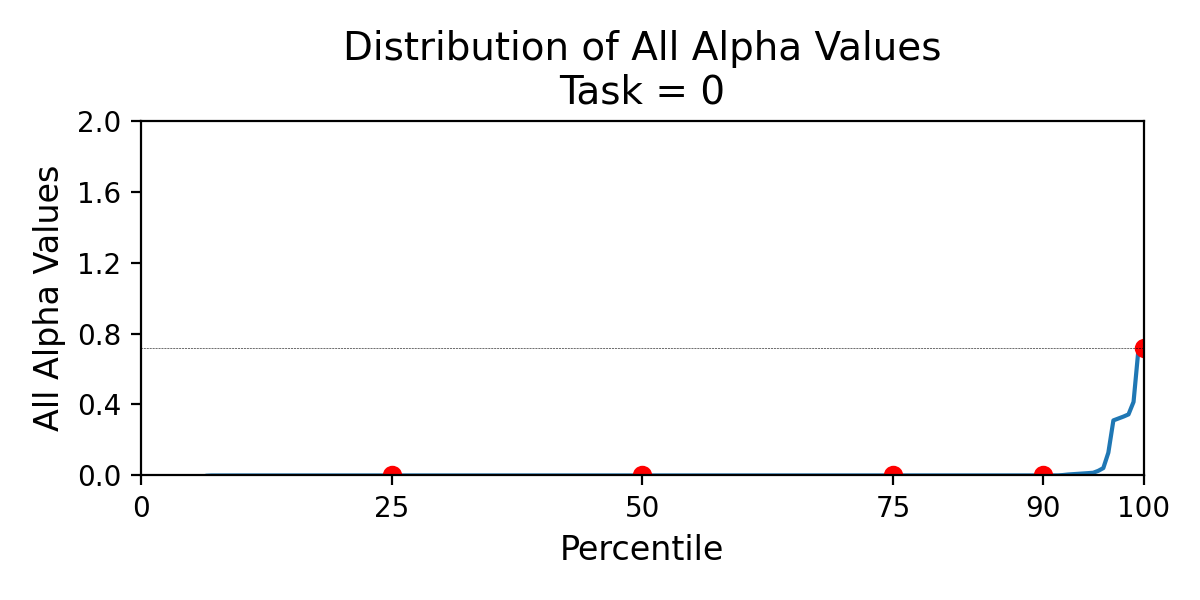}
        \vspace{-5mm}
        \caption{}
    \end{subfigure}
    \begin{subfigure}[t]{0.38\textwidth}
        \centering
        \includegraphics[trim=0 0 0 38, clip, width=\textwidth]{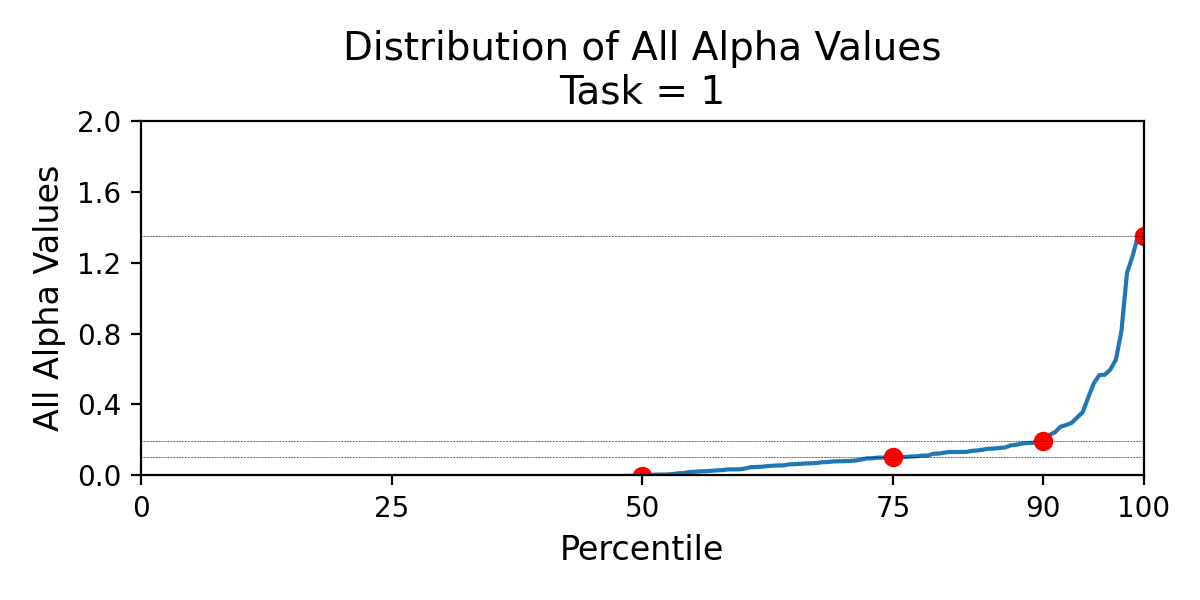}
        \vspace{-5mm}
        \caption{}
    \end{subfigure}
    \\
    \begin{subfigure}[t]{0.38\textwidth}
        \centering
        \vspace{2mm}
        \includegraphics[trim=0 0 0 38, clip, width=\textwidth]{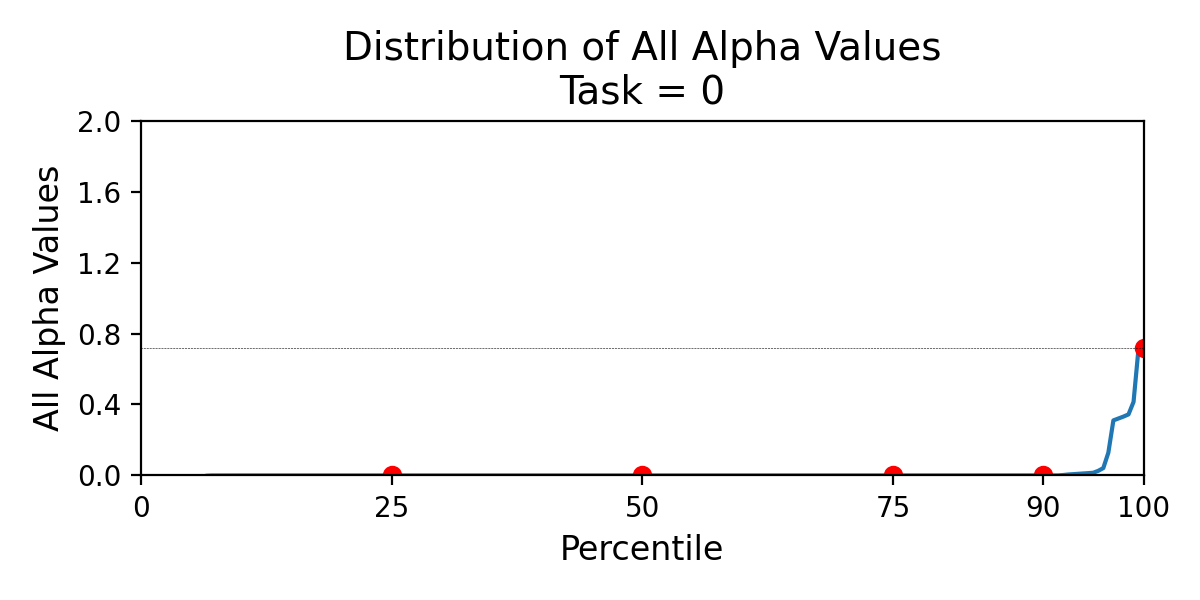}
        \vspace{-5mm}
        \caption{}
    \end{subfigure}
    \begin{subfigure}[t]{0.38\textwidth}
        \centering
        \vspace{2mm}
        \includegraphics[trim=0 0 0 38, clip, width=\textwidth]{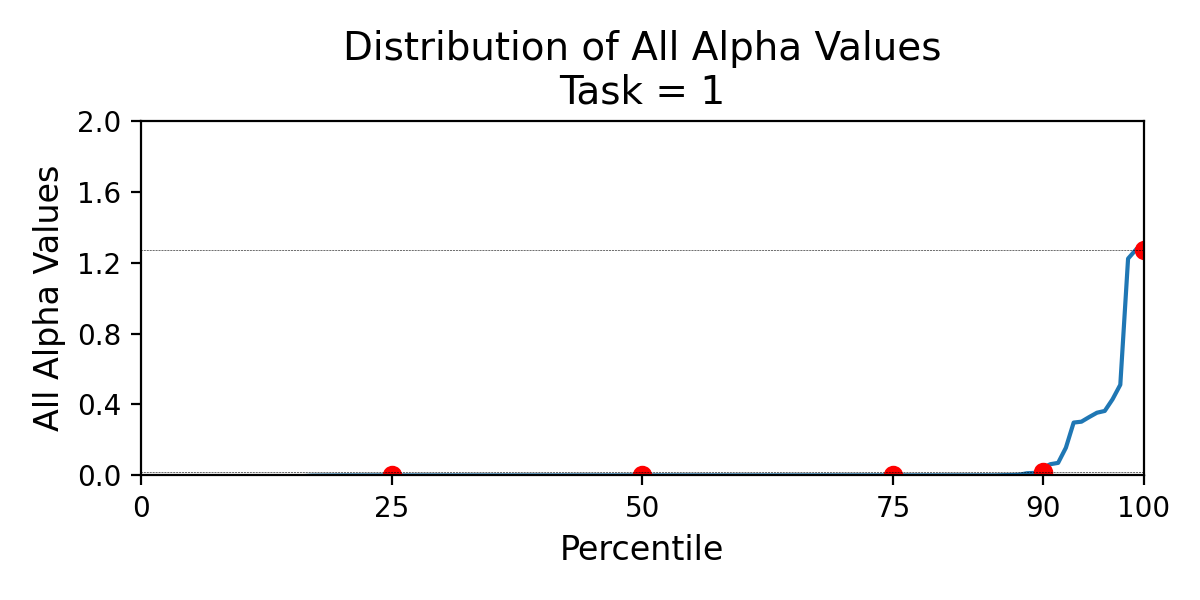}
        \vspace{-5mm}
        \caption{}
    \end{subfigure}
    \caption{
    \looseness=-1Distribution of $\alpha^{(i,j)}$ on ten task CIFAR-100 for (a) I-DARTS after training on task 1, (b) I-DARTS after training on task 2, (c) I-DARTS* after training on task 1, (d) I-DARTS* after training on task 2. Importantly, the distribution in (b) has changed rapidly from (a). We address this concern with an $\alpha$ regularization, and demonstrate that (d) is now much closer to (c).
    }
	\label{fig:alpha-motivation}
\end{figure*}

\section{Alpha Analysis and Regularization}

When analyzing the distribution of the $\alpha^{(i,j)}$ in our I-DARTS method, we made an interesting discovery. We found the $\alpha^{(i,j)}$ tend to increase after each task, which prevents useful learning in future tasks if the $\alpha^{(i,j)}$ are already converged to high values. We visualize this in Figure~\ref{fig:alpha-motivation}, showing that when training I-DARTS on two tasks of the ten task incremental CIFAR-100 benchmark, the distribution of $\alpha^{(i,j)}$ after training task 2 is of a very distinctly different distribution compared to after task 1. Because we can consider the model after task 1 to be a typical offline DARTS solution, the sharp distinction between (a) and (b) in Figure~\ref{fig:alpha-motivation} is alarming.

Using this finding, we propose to add an additional regularization to our learning objective to reduce this behavior in the $\alpha^{(i,j)}$ distribution. Our intuition is that $\alpha^{(i,j)}$ should be able to update in the future incremental learning tasks, and we therefore need to prevent over-fitting in the current task. We realize this intuition by adding an L2 distance regularization penalty on the $\alpha^{(i,j)}$. Specifically, we add the following regularization penalty:

\begin{equation}
    \mathcal{L}_{reg,\alpha}(\theta_n) = \sqrt{ \sum_{i,j \in \theta_n} \left ( \alpha^{(i,j)} \right) ^2}
    \label{eq:alpha}
\end{equation}

With this added loss function, we see in (c) and (d) of Figure~\ref{fig:alpha-motivation} that the distribution of $\alpha^{(i,j)}$ after task 2 is much closer to the distribution after task 1. We are now more confident that our super-model is learning \textit{stable} $\alpha^{(i,j)}$ over incremental tasks, which is reflected in our final results. We refer to I-DARTS with this regularization as I-DARTS*, with the following formal objective:

\begin{equation}
\begin{split}
    \underset{\theta_n}{min} \hspace{0.5em} \mathcal{L}_{CE} \left( p_{\theta_{n,1:n}}(x), y \right)  &+ \mu \mathcal{L}_{KD}(x,\theta_n,\theta_{n-1}) \\
    &+ \lambda \mathcal{L}_{reg,\alpha}(\theta_n)
\end{split}
\label{eq:lwf_alpha}
\end{equation}
where $\lambda$ is a scalar hyperparameter which weights the contribution of $\mathcal(L)_{reg,\alpha}$.

\section{Experiments}

We evaluate I-DARTS and I-DARTS* for an RF signal incremental classification benchmark and two incremental image classification benchmarks. We compare to several recent state-of-the-art approaches: elastic-weight consolidation (EWC)~\cite{kirkpatrick2017overcoming}, learning without forgetting (LwF)~\cite{li2016learning}, and end-to-end incremental learning (E2E)~\cite{castro2018end}. We also compare to a model trained with no incremental learning strategy (denoted as ``naive") and simple data replay (denoted as ``replay").

We emphasize that the impact of this work is not to incrementally push the state-of-the-art performance for a well-studied problem (such as incremental image classification), but rather to demonstrate that \textit{we can incrementally learn in applications where we must also simultaneously learn our architecture}. Thus, we benchmark with the \textit{RF classification dataset} to demonstrate the \textit{large performance gains that can be achieved on understudied modalities or datasets}. We then benchmark with common \textit{image classification datasets} to demonstrate that our method is \textit{comparable to state-of-the-art methods which use a highly developed architecture}.

\noindent
\textbf{Candidate Operations}: Following the original DARTS work, we include the following candidate operations in $\mathcal{O}$: $3 \times 3$ and $5 \times 5$ separable convolutions, $3 \times 3$ and $5 \times 5$ dilated separable convolutions, $3 \times 3$ max pooling, $3 \times 3$ average pooling, identity, and \textit{zero}. For the RF signal classification dataset, we replace the 2d convolutions in the DARTS operations with 1d convolution operations.

\noindent
\textbf{Evaluation Metrics}: Following prior works, we evaluate methods in the class-incremental learning setting using: (I) final performance, or the performance with respect to all past classes after having seen all $N$ tasks (referred to as $A_{N,1:N}$); (II) maximum model parameters, or the maximum model parameters at any given time (important for DARTS); (III) final model parameters, or the model parameters of the final inferred architecture; and (IV) the training time given in GPU days. We use index $k$ to index tasks through time and index $n$ to index tasks with respect to test/validation data (for example, $A_{k,n}$ describes the accuracy of our model after task $k$ on task $n$ data). Specifically:
\begin{equation}
    A_{k,n} = \frac{1}{|\mathcal{D}_n^{test}|} \sum_{(x,y)\in\mathcal{D}_n^{test}} \bm{1}(\hat{y}(x,\theta_{k,n}) = y \mid \hat{y} \in \mathcal{T}_n)
\end{equation}
For the final task accuracy in our results, we will denote $A_{N,1:N}$ as simply $A_{N}$.

\noindent
\textbf{Additional Experiment Details}: We augment image training data using standard augmentations such as random horizontal flips and crops. Our results were generated using 2080 Ti GPUs. In implementing our experiments, we leveraged both the Microsoft nni~\cite{nni} package for DARTS~\cite{liu2018darts} the Avalanche~\cite{lomonaco2021avalanche} continual learning library for incremental learning benchmarks.

We tuned hyperparameters using a grid search. The hyperparameters were tuned using k-fold cross validation with three folds of the training data on only half of the tasks. We do not tune hyperparameters on the full task set because tuning hyperparameters with hold out data from all tasks may violate the principle of continual learning that states each task in visited only once~\cite{Ven:2019}. The results reported are on testing splits (defined in the dataset).

\begin{table*}[h!]
\small
\caption{Ablation results for class-incremental learning on four-task Deepsig. Results are reported as an average of 2 runs. A coreset of 1000 samples is leveraged for each method.}
\centering
\begin{tabular}{c c | c c c c}
    \hline
    Method & Replay & $A_{N}$ (\%,$\uparrow$) & \thead{Max Model\\Params ($\downarrow$)}  & \thead{Final Model\\Params ($\downarrow$)} & GPU Days ($\downarrow$) \\
    \hline
    I-DARTS  & 1e3 & $56.3 \pm 0.0$ & $2.5e5$ & $5.7e4 \pm 4.6e3$ & $3.0$ \\ 
    I-DARTS* & 1e3 & $57.3 \pm 0.2$ & $2.5e5$ & $5.8e4 \pm 4.7e3$ & $3.0$ \\  
    \hline
    Remove Class Balancing  & 1e3 & $53.2 \pm 0.8$ & $2.5e5$ & $5.3e4 \pm 2.7e2$ & $3.0$ \\ 
    Remove KD & 1e3 & $56.1 \pm 0.0$ & $2.5e5$ & $4.8e4 \pm 5.3e3$ & $2.5$ \\ 
    Remove All (DARTS) & 1e3 & $55.0 \pm 0.0$ & $2.5e5$ & $5.6e4 \pm 2.0e3$ & $2.5$ \\ 
    \hline
\end{tabular}
\label{tab:deepsig-ab}
\end{table*}

\begin{table*}[h!]
\small
\caption{Results for class-incremental learning on four-task Deepsig. Results are reported as an average of 2 runs. A coreset of 1000 samples is leveraged for each replay-based method.}
\centering
\begin{tabular}{c c | c c c c}
    \hline
    Method & Replay & $A_{N}$ (\%,$\uparrow$) & \thead{Max Model\\Params ($\downarrow$)}  & \thead{Final Model\\Params ($\downarrow$)} & GPU Days ($\downarrow$) \\
    \hline
    Naive    & 0 & $18.5 \pm 0.5$ & $\bm{1.7e5}$ & $1.7e5$ & $4.6e-1$ \\ 
    EWC      & 0 & $18.3 \pm 0.0$ & $\bm{1.7e5}$ & $1.7e5$ & $4.9e-1$ \\ 
    LWF      & 0 & $24.4 \pm 1.6$ & $\bm{1.7e5}$ & $1.7e5$ & $4.9e-1$ \\ 
    Replay   & 1e3 & $44.1 \pm 0.7$ & $\bm{1.7e5}$ & $1.7e5$ & $4.7e-1$ \\ 
    E2E      & 1e3 & $47.3 \pm 0.6$ & $\bm{1.7e5}$ & $1.7e5$ & $5.4e-1$ \\ 
    \hline
    DARTS  & 1e3 & $55.0 \pm 0.0$ & $2.5e5$ & $\bm{5.6e4} \pm 2.0e3$ & $2.5$ \\ 
    I-DARTS  & 1e3 & $56.3 \pm 0.0$ & $2.5e5$ & $5.7e4 \pm 4.6e3$ & $3.0$ \\ 
    I-DARTS* & 1e3 & $\bm{57.3} \pm 0.2$ & $2.5e5$ & $5.8e4 \pm 4.7e3$ & $3.0$ \\ 
    \hline
\end{tabular}
\label{tab:deepsig}
\end{table*}

\begin{table*}[h!]
\small
\caption{Ablation results for class-incremental learning on ten-task CIFAR-100. Results are reported as an average of 2 runs. A coreset of 2000 images is leveraged for each method.}
\centering
\begin{tabular}{c c | c c c c}
    \hline
    Method & Replay & $A_{N}$ (\%,$\uparrow$) & \thead{Max Model\\Params ($\downarrow$)}  & \thead{Final Model\\Params ($\downarrow$)} & GPU Days ($\downarrow$) \\
    \hline
    I-DARTS  & 2e3 & $37.2 \pm 0.1$ & $2.0e6$ & $5.3e5 \pm 1.4e4$ & $6.4$ \\ 
    I-DARTS* & 2e3 & $37.7 \pm 0.1$ & $2.0e6$ & $4.8e5 \pm 1.3e4$ & $6.6$ \\  
    \hline
    Remove Class Balancing  & 2e3 & $36.7 \pm 0.4$ & $2.0e6$ & $5.0e5 \pm 5.7e4$ & $5.1$ \\ 
    Remove KD & 2e3 & $29..8 \pm 6.1$ & $2.0e6$ & $4.5e5 \pm 9.0e3$ & $5.1$ \\ 
    Remove All (DARTS) & 2e3 & $33.4 \pm 0.9$ & $2.0e6$ & $4.5e5 \pm 1.3e4$ & $3.9$ \\
    \hline
\end{tabular}
\label{tab:results_cifar100-ab}
\end{table*}

\begin{table*}[h!]
\small
\caption{Results for class-incremental learning on five-task CIFAR-10. Results are reported as an average of 2 runs. A coreset of 1000 images is leveraged for each replay-based method.}
\centering
\begin{tabular}{c c | c c c c}
    \hline
    Method & Replay & $A_{N}$ (\%,$\uparrow$) & \thead{Max Model\\Params ($\downarrow$)}  & \thead{Final Model\\Params ($\downarrow$)} & GPU Days ($\downarrow$) \\
    \hline
    Naive    & 0 & $19.7 \pm 0.0$ & $\bm{2.7e5}$ & $\bm{2.7e5}$ & $3.6e-1$ \\ 
    EWC      & 0 & $19.6 \pm 0.0$ & $\bm{2.7e5}$ & $\bm{2.7e5}$ & $4.5e-1$ \\ 
    LWF      & 0 & $22.0 \pm 0.5$ & $\bm{2.7e5}$ & $\bm{2.7e5}$ & $4.2e-1$ \\ 
    Replay   & 1e3 & $60.8 \pm 0.8$ & $\bm{2.7e5}$ & $\bm{2.7e5}$ & $3.9e-1$ \\ 
    E2E      & 1e3 & $\bm{73.5} \pm 0.8$ & $\bm{2.7e5}$ & $\bm{2.7e5}$ & $5.0e-1$ \\ 
    \hline
    DARTS  & 1e3 & $64.6 \pm 0.1$ & $1.95e6$ & $3.1e5 \pm 3.1e3$ & $3.9$ \\
    I-DARTS  & 1e3 & $71.3 \pm 0.1$ & $1.95e6$ & $3.8e5 \pm 3.2e3$ & $4.9$ \\ 
    I-DARTS* & 1e3 & $70.6 \pm 5.0$ & $1.95e6$ & $4.0e5 \pm 3.5e3$ & $4.3$ \\ 
    \hline
\end{tabular}
\label{tab:results_cifar10}
\end{table*}

\begin{table*}[h!]
\small
\caption{Results for class-incremental learning on ten-task CIFAR-100. Results are reported as an average of 2 runs. A coreset of 2000 images is leveraged for each replay-based method.}
\centering
\begin{tabular}{c c | c c c c}
    \hline
    Method & Replay & $A_{N}$ (\%,$\uparrow$) & \thead{Max Model\\Params ($\downarrow$)}  & \thead{Final Model\\Params ($\downarrow$)} & GPU Days ($\downarrow$) \\
    \hline
    Naive    & 0 & $8.9 \pm 0.0$ & $\bm{2.8e5}$ & $\bm{2.8e5}$ & $2.9e-1$ \\ 
    EWC      & 0 & $9.0 \pm 0.0$ & $\bm{2.8e5}$ & $\bm{2.8e5}$ & $3.7e-1$ \\ 
    LWF      & 0 & $16.7 \pm 0.1$ & $\bm{2.8e5}$ & $\bm{2.8e5}$ & $3.4e-1$ \\ 
    Replay   & 2e3 & $36.2 \pm 0.2$ & $\bm{2.8e5}$ & $\bm{2.8e5}$ & $3.7e-1$ \\ 
    E2E      & 2e3 & $\bm{41.9} \pm 0.6$ & $\bm{2.8e5}$ & $\bm{2.8e5}$ & $5.3e-1$ \\ 
    \hline
    DARTS  & 2e3 & $33.4 \pm 0.9$ & $2.0e6$ & $4.5e5 \pm 1.3e4$ & $3.9$ \\
    I-DARTS  & 2e3 & $37.2 \pm 0.1$ & $2.0e6$ & $5.3e5 \pm 1.4e4$ & $6.4$ \\ 
    I-DARTS* & 2e3 & $37.7 \pm 0.1$ & $2.0e6$ & $4.8e5 \pm 1.3e4$ & $6.6$ \\ 
    \hline
\end{tabular}
\label{tab:results_cifar100}
\end{table*}

\subsection{I-DARTS is SOTA for RF signal classification}

Our first benchmark is four-task incremental learning on the RADIOML 2018.01a DEEPSIG dataset~\cite{o2018over} for RF signal classification. The DEEPSIG dataset includes 2 million examples, each 1024 samples long, which have been modulated by 24 digital and analog modulation types, and the classification task is to predict which of the 24 modulation types was used for each signal. This benchmark exemplifies the key contribution of our work because there is no clear neural architecture to apply on this dataset.

We borrowed the RF ResNet~\cite{he2016deep} network introduced in the original DEEPSIG work~\cite{o2018over}. This model contains 6 Residual stacks followed by three fully connected layers. We also tested the second architecture listed in the work, but do not report results given they far under-perform the ResNet variant. For the RADIOML 2018.01a DEEPSIG dataset~\cite{o2018over}, we created incremental learning tasks by allocating modulation types such that similar modulation types appeared in the same task. The motivation behind this is to make the tasks more ``realistic". We reemphasize that this benchmark exemplifies the key contribution of our work because there is no clear neural architecture to apply on this dataset.

We train our super-model using the DARTS bi-level optimization for 50 epochs with the Adam optimzer and a learning rate of 0.05 for the architecture weights and 5e-3 for the $\alpha$ weights. For the second training stage, we train for 125 epochs. The learning rate is set to 0.05 and is reduced by 10 after 50, 75, and 100 epochs. We use a weight decay of 0.0002 and batch size of 128. Using a simple grid search to find hyperparameters for Eq~\eqref{eq:lwf_alpha}, we found $\mu$ to be 0.5 and $\lambda$ to be 1e-3.

We first compare I-DARTS and I-DARTS* to regular DARTS with stored replay data, as well as two additional ablations on our method, in Table~\ref{tab:deepsig-ab}. We see that each of our contributions incrementally improves the DARTS performance without sacrificing the number of model parameters or training time in GPU days.

In Table~\ref{tab:deepsig}, we compare out approach to several popular incremental learning methods (listed in the beginning of this section). We implemented all of these methods with the two different neural architectures from the original paper~\cite{o2018over}, and only report the higher performing architecture, which is build on Residual blocks~\cite{he2016deep}. \textit{We show here that our approach makes substantial improvements in accuracy over the existing methods, while having both a similar maximum number of architecture parameters and a smaller number of final model parameters}. We observe that this performance comes at a cost in GPU days, but we argue this is less important given that the significant engineering time likely went into designing the baseline model, which is not accounted for in our metrics.

\subsection{I-DARTS is near SOTA for image classification}

Next, we benchmark our method using the CIFAR-10 and CIFAR-100 datasets~\cite{krizhevsky2009learning} class-incremental learning benchmarks. These datasets contains 10/100 classes of 32x32x3 images, respectively. We only benchmark image classification on the two CIFAR benchmarks (and not the ImageNet benchmarks) because DARTS on ImagenNet~\cite{russakovsky2015imagenet} is beyond our computational limits. Following prior work~\cite{wu2019large}, we train with a 32-layer ResNet~\cite{he2016deep}. We use the same training protocol from the RF classification experiments.

Similar to the RF classification experiments, we first compare I-DARTS and I-DARTS* to regular DARTS with stored replay data, as well as two additional ablations on our method (given in Table~\ref{tab:results_cifar100-ab}). We again see that each of our contributions incrementally improves the DARTS performance without sacrificing the number of model parameters or training time in GPU days.

We see a different story in the results from Tables~\ref{tab:results_cifar10}/\ref{tab:results_cifar100}, which compares performance to the same incremental learning methods. We actually find that E2E~\cite{castro2018end} achieves the highest performance, with I-DARTS* closely behind. \textit{Why does our method not improve CIFAR image classification?} We conjecture that there is actually a simple and intuitive explanation here. The ResNet~\cite{he2016deep} architecture has been profoundly studied for this dataset, with the optimal architecture settings evolving over years of research. In fact, most of the incremental learning we list here were developed specifically for this architecture. Keeping this in mind, it makes sense that there is little to none performance gap to be closed with our incremental DARTS methods. \textit{Rather, the story here is that our contributions improve DARTS in the incremental setting for this image classification dataset}.
\section{Conclusions}

We unify two concepts crucial for machine learning applications: \textit{incremental learning} and \textit{neural architecture search}. Our motivation is to build reliable models for applications which undergo \textit{continuous domain shifts} in training data after deployment, while being flexible to handle data of \textit{less explored modalities} or \textit{unknown underlying complexity}. We start by proposing a baseline method, I-DARTS, which unifies concepts from (i) the one-shot NAS approach DARTS and (ii)  state-of-the-art incremental learning methods. We analyze the performance of I-DARTS and use this to motivate regularization on the DARTS mixture weights, further boosting our performance. We demonstrate that I-DARTS achieves high performance gains on an RF classification task in which the existing neural architecture has not been highly engineered, while additionally demonstrating that we are highly competitive on state-of-the-art image classification tasks. The main takeaway from our work is that \textit{our method can successfully learn both a neural architecture and architecture weights in simultaneous fashion, enabling new real-world application for machine learning models}.

\section*{Acknowledgements}

This material is based upon work supported by DARPA under Contract No. N6523620C8020. Any opinions, findings, and conclusions or recommendations expressed in this material are those of the author(s) and do not necessarily reflect the views of the US government/ Department of Defense.

\bibliographystyle{IEEEtran}
\bibliography{references}

\begin{thebibliography}{10}
\providecommand{\url}[1]{#1}
\csname url@samestyle\endcsname
\providecommand{\newblock}{\relax}
\providecommand{\bibinfo}[2]{#2}
\providecommand{\BIBentrySTDinterwordspacing}{\spaceskip=0pt\relax}
\providecommand{\BIBentryALTinterwordstretchfactor}{4}
\providecommand{\BIBentryALTinterwordspacing}{\spaceskip=\fontdimen2\font plus
\BIBentryALTinterwordstretchfactor\fontdimen3\font minus
  \fontdimen4\font\relax}
\providecommand{\BIBforeignlanguage}[2]{{%
\expandafter\ifx\csname l@#1\endcsname\relax
\typeout{** WARNING: IEEEtran.bst: No hyphenation pattern has been}%
\typeout{** loaded for the language `#1'. Using the pattern for}%
\typeout{** the default language instead.}%
\else
\language=\csname l@#1\endcsname
\fi
#2}}
\providecommand{\BIBdecl}{\relax}
\BIBdecl

\bibitem{goodfellow2013empirical}
I.~J. Goodfellow, M.~Mirza, D.~Xiao, A.~Courville, and Y.~Bengio, ``An
  empirical investigation of catastrophic forgetting in gradient-based neural
  networks,'' \emph{arXiv preprint arXiv:1312.6211}, 2013.

\bibitem{he2016deep}
K.~He, X.~Zhang, S.~Ren, and J.~Sun, ``Deep residual learning for image
  recognition,'' in \emph{Proceedings of the IEEE conference on computer vision
  and pattern recognition}, 2016, pp. 770--778.

\bibitem{russakovsky2015imagenet}
O.~Russakovsky, J.~Deng, H.~Su, J.~Krause, S.~Satheesh, S.~Ma, Z.~Huang,
  A.~Karpathy, A.~Khosla, M.~Bernstein \emph{et~al.}, ``Imagenet large scale
  visual recognition challenge,'' \emph{International journal of computer
  vision}, vol. 115, no.~3, pp. 211--252, 2015.

\bibitem{o2018over}
T.~J. O’Shea, T.~Roy, and T.~C. Clancy, ``Over-the-air deep learning based
  radio signal classification,'' \emph{IEEE Journal of Selected Topics in
  Signal Processing}, vol.~12, no.~1, pp. 168--179, 2018.

\bibitem{castro2018end}
F.~M. Castro, M.~J. Mar{\'\i}n-Jim{\'e}nez, N.~Guil, C.~Schmid, and K.~Alahari,
  ``End-to-end incremental learning,'' in \emph{Proceedings of the European
  Conference on Computer Vision (ECCV)}, 2018, pp. 233--248.

\bibitem{ebrahimi2020adversarial}
S.~Ebrahimi, F.~Meier, R.~Calandra, T.~Darrell, and M.~Rohrbach, ``Adversarial
  continual learning,'' \emph{arXiv preprint arXiv:2003.09553}, 2020.

\bibitem{lee2020neural}
S.~Lee, J.~Ha, D.~Zhang, and G.~Kim, ``A neural dirichlet process mixture model
  for task-free continual learning,'' \emph{arXiv preprint arXiv:2001.00689},
  2020.

\bibitem{lomonaco2017core50}
V.~Lomonaco and D.~Maltoni, ``Core50: a new dataset and benchmark for
  continuous object recognition,'' \emph{arXiv preprint arXiv:1705.03550},
  2017.

\bibitem{maltoni2019continuous}
D.~Maltoni and V.~Lomonaco, ``Continuous learning in single-incremental-task
  scenarios,'' \emph{Neural Networks}, vol. 116, pp. 56--73, 2019.

\bibitem{Rusu:2016}
A.~A. Rusu, N.~C. Rabinowitz, G.~Desjardins, H.~Soyer, J.~Kirkpatrick,
  K.~Kavukcuoglu, R.~Pascanu, and R.~Hadsell, ``Progressive neural networks,''
  \emph{arXiv preprint arXiv:1606.04671}, 2016.

\bibitem{aljundi2017memory}
R.~Aljundi, F.~Babiloni, M.~Elhoseiny, M.~Rohrbach, and T.~Tuytelaars, ``Memory
  aware synapses: Learning what (not) to forget,'' in \emph{ECCV}, 2018.

\bibitem{ebrahimi2019uncertainty}
S.~Ebrahimi, M.~Elhoseiny, T.~Darrell, and M.~Rohrbach, ``Uncertainty-guided
  continual learning with bayesian neural networks,'' \emph{arXiv preprint
  arXiv:1906.02425}, 2019.

\bibitem{kirkpatrick2017overcoming}
J.~Kirkpatrick, R.~Pascanu, N.~Rabinowitz, J.~Veness, G.~Desjardins, A.~A.
  Rusu, K.~Milan, J.~Quan, T.~Ramalho, A.~Grabska-Barwinska \emph{et~al.},
  ``Overcoming catastrophic forgetting in neural networks,'' \emph{Proceedings
  of the national academy of sciences}, 2017.

\bibitem{titsias2019functional}
M.~K. Titsias, J.~Schwarz, A.~G. d.~G. Matthews, R.~Pascanu, and Y.~W. Teh,
  ``Functional regularisation for continual learning with gaussian processes,''
  in \emph{International Conference on Learning Representations}, 2019.

\bibitem{zenke2017continual}
F.~Zenke, B.~Poole, and S.~Ganguli, ``Continual learning through synaptic
  intelligence,'' in \emph{International Conference on Machine Learning}, 2017.

\bibitem{hou2018lifelong}
S.~Hou, X.~Pan, C.~Change~Loy, Z.~Wang, and D.~Lin, ``Lifelong learning via
  progressive distillation and retrospection,'' in \emph{Proceedings of the
  European Conference on Computer Vision (ECCV)}, 2018, pp. 437--452.

\bibitem{lee2019overcoming}
K.~Lee, K.~Lee, J.~Shin, and H.~Lee, ``Overcoming catastrophic forgetting with
  unlabeled data in the wild,'' in \emph{Proceedings of the IEEE International
  Conference on Computer Vision}, 2019, pp. 312--321.

\bibitem{li2016learning}
Z.~Li and D.~Hoiem, ``Learning without forgetting,'' \emph{IEEE transactions on
  pattern analysis and machine intelligence}, vol.~40, no.~12, pp. 2935--2947,
  2017.

\bibitem{elsken2019neural}
T.~Elsken, J.~H. Metzen, and F.~Hutter, ``Neural architecture search: A
  survey,'' \emph{The Journal of Machine Learning Research}, vol.~20, no.~1,
  pp. 1997--2017, 2019.

\bibitem{ren2020comprehensive}
P.~Ren, Y.~Xiao, X.~Chang, P.-Y. Huang, Z.~Li, X.~Chen, and X.~Wang, ``A
  comprehensive survey of neural architecture search: Challenges and
  solutions,'' \emph{arXiv preprint arXiv:2006.02903}, 2020.

\bibitem{wistuba2019survey}
M.~Wistuba, A.~Rawat, and T.~Pedapati, ``A survey on neural architecture
  search,'' \emph{arXiv preprint arXiv:1905.01392}, 2019.

\bibitem{cai2018proxylessnas}
H.~Cai, L.~Zhu, and S.~Han, ``Proxylessnas: Direct neural architecture search
  on target task and hardware,'' \emph{arXiv preprint arXiv:1812.00332}, 2018.

\bibitem{guo2020single}
Z.~Guo, X.~Zhang, H.~Mu, W.~Heng, Z.~Liu, Y.~Wei, and J.~Sun, ``Single path
  one-shot neural architecture search with uniform sampling,'' in
  \emph{European Conference on Computer Vision}.\hskip 1em plus 0.5em minus
  0.4em\relax Springer, 2020, pp. 544--560.

\bibitem{pham2018efficient}
H.~Pham, M.~Guan, B.~Zoph, Q.~Le, and J.~Dean, ``Efficient neural architecture
  search via parameters sharing,'' in \emph{International Conference on Machine
  Learning}.\hskip 1em plus 0.5em minus 0.4em\relax PMLR, 2018, pp. 4095--4104.

\bibitem{wu2019fbnet}
B.~Wu, X.~Dai, P.~Zhang, Y.~Wang, F.~Sun, Y.~Wu, Y.~Tian, P.~Vajda, Y.~Jia, and
  K.~Keutzer, ``Fbnet: Hardware-aware efficient convnet design via
  differentiable neural architecture search,'' in \emph{Proceedings of the
  IEEE/CVF Conference on Computer Vision and Pattern Recognition}, 2019, pp.
  10\,734--10\,742.

\bibitem{liu2018darts}
H.~Liu, K.~Simonyan, and Y.~Yang, ``Darts: Differentiable architecture
  search,'' \emph{arXiv preprint arXiv:1806.09055}, 2018.

\bibitem{nni}
Microsoft, ``Neural network intelligence,'' 2020.

\bibitem{wu2021firefly}
L.~Wu, B.~Liu, P.~Stone, and Q.~Liu, ``Firefly neural architecture descent: a
  general approach for growing neural networks,'' \emph{arXiv preprint
  arXiv:2102.08574}, 2021.

\bibitem{li2019learn}
X.~Li, Y.~Zhou, T.~Wu, R.~Socher, and C.~Xiong, ``Learn to grow: A continual
  structure learning framework for overcoming catastrophic forgetting,'' in
  \emph{International Conference on Machine Learning}.\hskip 1em plus 0.5em
  minus 0.4em\relax PMLR, 2019, pp. 3925--3934.

\bibitem{zhang2020one}
M.~Zhang, H.~Li, S.~Pan, X.~Chang, C.~Zhou, Z.~Ge, and S.~W. Su, ``One-shot
  neural architecture search: Maximising diversity to overcome catastrophic
  forgetting,'' \emph{IEEE Transactions on Pattern Analysis and Machine
  Intelligence}, 2020.

\bibitem{huang2019neural}
S.~Huang, V.~Fran{\c{c}}ois-Lavet, and G.~Rabusseau, ``Neural architecture
  search for class-incremental learning,'' \emph{arXiv preprint
  arXiv:1909.06686}, 2019.

\bibitem{mundt2021neural}
M.~Mundt, I.~Pliushch, and V.~Ramesh, ``Neural architecture search of deep
  priors: Towards continual learning without catastrophic interference,'' in
  \emph{Proceedings of the IEEE/CVF Conference on Computer Vision and Pattern
  Recognition}, 2021, pp. 3523--3532.

\bibitem{hsu2018re}
Y.-C. Hsu, Y.-C. Liu, A.~Ramasamy, and Z.~Kira, ``Re-evaluating continual
  learning scenarios: A categorization and case for strong baselines,''
  \emph{arXiv preprint arXiv:1810.12488}, 2018.

\bibitem{aljundi2019online}
R.~Aljundi, E.~Belilovsky, T.~Tuytelaars, L.~Charlin, M.~Caccia, M.~Lin, and
  L.~Page-Caccia, ``Online continual learning with maximal interfered
  retrieval,'' in \emph{Advances in Neural Information Processing Systems},
  2019, pp. 11\,849--11\,860.

\bibitem{aljundi2019gradient}
R.~Aljundi, M.~Lin, B.~Goujaud, and Y.~Bengio, ``Gradient based sample
  selection for online continual learning,'' in \emph{Advances in Neural
  Information Processing Systems}, 2019, pp. 11\,816--11\,825.

\bibitem{chaudhry2018efficient}
A.~Chaudhry, M.~Ranzato, M.~Rohrbach, and M.~Elhoseiny, ``Efficient lifelong
  learning with a-{GEM},'' in \emph{International Conference on Learning
  Representations}, 2019.

\bibitem{chaudhry2019episodic}
A.~Chaudhry, M.~Rohrbach, M.~Elhoseiny, T.~Ajanthan, P.~K. Dokania, P.~H. Torr,
  and M.~Ranzato, ``Continual learning with tiny episodic memories,''
  \emph{arXiv preprint arXiv:1902.10486}, 2019.

\bibitem{Gepperth:2017}
A.~Gepperth and C.~Karaoguz, ``Incremental learning with self-organizing
  maps,'' \emph{2017 12th International Workshop on Self-Organizing Maps and
  Learning Vector Quantization, Clustering and Data Visualization (WSOM)}, pp.
  1--8, 2017.

\bibitem{hayes2018memory}
T.~L. Hayes, N.~D. Cahill, and C.~Kanan, ``Memory efficient experience replay
  for streaming learning,'' in \emph{2019 International Conference on Robotics
  and Automation (ICRA)}.\hskip 1em plus 0.5em minus 0.4em\relax IEEE, 2019,
  pp. 9769--9776.

\bibitem{hou2019learning}
S.~Hou, X.~Pan, C.~C. Loy, Z.~Wang, and D.~Lin, ``Learning a unified classifier
  incrementally via rebalancing,'' in \emph{Proceedings of the IEEE Conference
  on Computer Vision and Pattern Recognition}, 2019, pp. 831--839.

\bibitem{Kemker:2017}
R.~Kemker, M.~McClure, A.~Abitino, T.~Hayes, and C.~Kanan, ``Measuring
  catastrophic forgetting in neural networks,'' \emph{AAAI Conference on
  Artificial Intelligence}, 2018.

\bibitem{Lopez-Paz:2017}
D.~Lopez-Paz and M.~Ranzato, ``Gradient episodic memory for continual
  learning,'' in \emph{Proceedings of the 31st International Conference on
  Neural Information Processing Systems}, ser. NIPS'17.\hskip 1em plus 0.5em
  minus 0.4em\relax USA: Curran Associates Inc., 2017, pp. 6470--6479.

\bibitem{Rebuffi:2016}
S.~Rebuffi, A.~Kolesnikov, G.~Sperl, and C.~H. Lampert, ``icarl: Incremental
  classifier and representation learning,'' in \emph{2017 {IEEE} Conference on
  Computer Vision and Pattern Recognition}, ser. CVPR'17, 2017, pp. 5533--5542.

\bibitem{robins1995catastrophic}
A.~Robins, ``Catastrophic forgetting, rehearsal and pseudorehearsal,''
  \emph{Connection Science}, vol.~7, no.~2, pp. 123--146, 1995.

\bibitem{rolnick2019experience}
D.~Rolnick, A.~Ahuja, J.~Schwarz, T.~Lillicrap, and G.~Wayne, ``Experience
  replay for continual learning,'' in \emph{Advances in Neural Information
  Processing Systems}, 2019, pp. 348--358.

\bibitem{von2019continual}
J.~von Oswald, C.~Henning, J.~Sacramento, and B.~F. Grewe, ``Continual learning
  with hypernetworks,'' \emph{arXiv preprint arXiv:1906.00695}, 2019.

\bibitem{lesort2019generative}
T.~Lesort, H.~Caselles-Dupr{\'e}, M.~Garcia-Ortiz, A.~Stoian, and D.~Filliat,
  ``Generative models from the perspective of continual learning,'' in
  \emph{2019 International Joint Conference on Neural Networks (IJCNN)}.\hskip
  1em plus 0.5em minus 0.4em\relax IEEE, 2019, pp. 1--8.

\bibitem{van2018generative}
G.~M. van~de Ven and A.~S. Tolias, ``Generative replay with feedback
  connections as a general strategy for continual learning,'' \emph{arXiv
  preprint arXiv:1809.10635}, 2018.

\bibitem{welling2009herding}
M.~Welling, ``Herding dynamical weights to learn,'' in \emph{Proceedings of the
  26th Annual International Conference on Machine Learning}, 2009, pp.
  1121--1128.

\bibitem{lomonaco2021avalanche}
V.~Lomonaco, L.~Pellegrini, A.~Cossu, A.~Carta, G.~Graffieti, T.~L. Hayes,
  M.~De~Lange, M.~Masana, J.~Pomponi, G.~M. van~de Ven \emph{et~al.},
  ``Avalanche: an end-to-end library for continual learning,'' in
  \emph{Proceedings of the IEEE/CVF Conference on Computer Vision and Pattern
  Recognition}, 2021, pp. 3600--3610.

\bibitem{Ven:2019}
G.~M. van~de Ven and A.~S. Tolias, ``Three scenarios for continual learning,''
  \emph{arXiv preprint arXiv:1904.07734}, 2019.

\bibitem{krizhevsky2009learning}
A.~Krizhevsky, G.~Hinton \emph{et~al.}, ``Learning multiple layers of features
  from tiny images,'' \emph{Tech Report}, 2009.

\bibitem{wu2019large}
Y.~Wu, Y.~Chen, L.~Wang, Y.~Ye, Z.~Liu, Y.~Guo, and Y.~Fu, ``Large scale
  incremental learning,'' in \emph{Proceedings of the IEEE/CVF Conference on
  Computer Vision and Pattern Recognition}, 2019, pp. 374--382.

\end{thebibliography}

\end{document}